\documentclass[11pt]{article}
\usepackage[utf8]{inputenc}
\usepackage{amsmath,amssymb,amsfonts}
\usepackage{graphicx}
\usepackage{hyperref}
\usepackage{xcolor}
\usepackage{geometry}
\title{The Axiom-Based Atlas: A Structural Mapping of Theorems via Foundational Proof Vectors}
\author{Harim Yoo \\ Texas A\&M University}
\date{\today}

\begin{document}

\maketitle

\begin{abstract}
The Axiom-Based Atlas is a novel framework that structurally represents mathematical theorems as proof vectors over foundational axiom systems. By mapping the logical dependencies of theorems onto vectors indexed by axioms---such as those from Hilbert geometry, Peano arithmetic, or ZFC---we offer a new way to visualize, compare, and analyze mathematical knowledge. This vector-based formalism not only captures the logical foundation of theorems but also enables quantitative similarity metrics---such as cosine distance---between mathematical results, offering a new analytic layer for structural comparison. Using heatmaps, vector clustering, and AI-assisted modeling, this atlas enables the grouping of theorems by logical structure, not just by mathematical domain. We also introduce a prototype assistant (Atlas-GPT) that interprets natural language theorems and suggests likely proof vectors, supporting future applications in automated reasoning, mathematical education, and formal verification.

This direction is partially inspired by Terence Tao’s recent reflections on the convergence of symbolic and structural mathematics. The Axiom-Based Atlas aims to provide a scalable, interpretable model of mathematical reasoning that is both human-readable and AI-compatible, contributing to the future landscape of formal mathematical systems.
\end{abstract}

\section{Introduction}

Mathematics has long been structured around the formulation and proof of theorems from foundational axioms. While the logical rigor of formal proofs has been deeply explored through systems such as Hilbert-style axiomatizations and modern proof assistants, the structural comparison between theorems---in terms of their logical foundations---has remained elusive. Traditionally, theorems are classified by domain (e.g., geometry, number theory, algebra), but this categorization overlooks the shared logical underpinnings that transcend subject boundaries.\\

The Axiom-Based Atlas aims to address this gap by representing mathematical theorems as vectors over axiom systems. Each component of a vector corresponds to an axiom from a particular formal system, and the presence or absence of that axiom in the proof is captured as a binary or weighted value. This representation allows us to visualize, compare, and analyze the logical structure of theorems across different domains.\\

This project emerged from a desire to create a unified language for the foundational footprint of mathematical knowledge. Drawing inspiration from Gödel's encoding of formal systems and modern developments in vector representation, we propose a framework where logical dependencies are not only tracked but also quantified. As a result, we can define similarity metrics---such as cosine distance---between theorems, cluster them into logical families, and explore structural relationships that were previously implicit.\\

The Axiom-Based Atlas provides not only a static documentation of mathematical foundations but also a dynamic interface to engage with them. It is designed to support future integration with AI systems, including language models that can interpret natural language theorems and generate likely axiom vectors. Our prototype, Atlas-GPT, takes a step in this direction by offering semantic parsing, vector prediction, and structural explanation.\\

This framework contributes to a growing trend in mathematics where formalism, structure, and computation are increasingly unified. Inspired by Terence Tao's vision of mathematics as a navigable landscape of ideas, we believe the Axiom-Based Atlas opens a new path for exploring, organizing, and interacting with the foundational terrain of mathematics.

\section{Motivation and Related Work}

The motivation for the Axiom-Based Atlas stems from a desire to bridge the gap between formal mathematical logic and the human process of understanding and organizing mathematical knowledge. While formal systems provide a rigorous foundation for mathematics, they often remain disconnected from how mathematicians and students conceptualize the structure and interdependence of theorems.\\

Mathematical structuralism has long emphasized the relationships between objects rather than the objects themselves. This viewpoint aligns with the idea that theorems can be understood and classified based on the axioms they depend on, rather than solely their content or field. Our work proposes a concrete realization of this perspective by encoding the logical dependencies of theorems as vectors over a fixed axiom basis.\\

Several influential thinkers have anticipated aspects of this vision. The foundational work of David Hilbert emphasized the axiomatic method as a universal framework for mathematical thought. Kurt Gödel introduced the notion of encoding symbolic logic as numbers, demonstrating that metamathematical structure could be studied internally within a system. More recently, Terence Tao has explored how mathematics might evolve into a more dynamic, data-driven landscape, where ideas are navigated structurally, and connections between fields are made explicit.\\

In parallel, developments in automated theorem proving and proof assistants such as Lean, Coq, and Isabelle have enabled the formal verification of complex mathematical statements. While these tools focus on correctness and construction, our atlas is complementary: it emphasizes the structural pattern that emerges across proofs, abstracting away from full formalization to provide an interpretable model.\\

We also build upon recent work in AI-assisted mathematics, particularly in natural language theorem parsing, mathematical embedding spaces, and machine learning models for proof prediction. The Axiom-Based Atlas integrates these advances into a coherent framework that not only visualizes logic but quantifies it, enabling both human and machine interaction with the foundational architecture of mathematics.

\section{Methodology: Axioms and Proof Vectors}

The central idea of the Axiom-Based Atlas is to represent each mathematical theorem as a vector whose components correspond to the axioms used in its proof. This structure, called a \textit{proof vector}, allows for the systematic encoding, comparison, and visualization of the logical foundations of mathematical results.

\subsection*{Axiom Systems as Bases}
We begin by selecting formal axiom systems as the coordinate bases. In this study, we include well-known foundational systems such as
\begin{itemize}
  \item Hilbert's Geometry Axioms
  \item Peano Arithmetic
  \item Zermelo-Fraenkel Set Theory with Choice (ZFC)
  \item Vector Space Axioms
  \item Group Theory Axioms
\end{itemize}
Each axiom system is ordered and indexed so that every axiom occupies a fixed coordinate position in the vector space. For example, Hilbert’s Geometry Axioms may span a 12-dimensional vector, with entries representing individual axioms from incidence, order, congruence, and continuity groups. More axioms can be added to the Axiom-Based Atlas.

\subsection*{Proof Vectors}
Given a theorem, its proof vector is constructed by marking which axioms are invoked in its derivation. We primarily use binary vectors—each entry is 1 if the axiom is used, and 0 otherwise. In some contexts, we may also employ weighted values to reflect intensity, depth, or frequency of use.\\

This simple representation opens the door to a range of analytic techniques. Proof vectors can be
\begin{itemize}
  \item Visualized via heatmaps and similarity matrices
  \item Clustered into logical families using vector norms and distance metrics
  \item Compared using cosine similarity, Euclidean distance, or Jaccard index
\end{itemize}

\subsection*{Axiomatic Scope and Modularity}
By separating theorems by axiom system, we preserve modularity and avoid conflating logically unrelated results. At the same time, the framework supports the expansion to other systems (e.g., ring theory, topology, lambda calculus) and cross-system vector concatenation for composite theories.

\subsection*{Interoperability with AI Models}
Because the vectors are fixed-length, interpretable, and aligned with formal foundations, they can serve as training inputs, feature embeddings, or target labels in AI models. This paves the way for systems that can both learn from and reason about mathematical structure using foundational logic as a scaffold.\\

In this way, the Axiom-Based Atlas provides a minimal yet expressive representation of formal reasoning---one that is robust enough for machine interpretation and transparent enough for human insight.

\section{Theorem Examples and Vectors}

To demonstrate the practical utility of the Axiom-Based Atlas, we present a set of representative theorems across different axiom systems, each paired with its corresponding proof vector. These vectors reflect which axioms are employed in the proof of the theorem. The examples illustrate how logically distinct results can be embedded in a common vector space, opening the door to structural comparison.

\subsection*{Hilbert Geometry Axioms (12 dimensions)}
\begin{center}
\begin{tabular}{|l|l|}
\hline
\textbf{Theorem} & \textbf{Proof Vector} \\
\hline
Pythagorean Theorem & [1, 1, 1, 1, 1, 0, 1, 1, 1, 1, 1, 1] \\
Sum of Angles in Triangle & [1, 1, 1, 1, 1, 1, 1, 0, 1, 1, 1, 1] \\
Euler Line (Orthocenter–Centroid–Circumcenter) & [1, 1, 1, 1, 1, 0, 1, 1, 1, 1, 0, 1] \\
\hline
\end{tabular}
\end{center}

\subsection*{Peano Arithmetic (5 dimensions)}
\begin{center}
\begin{tabular}{|l|l|}
\hline
\textbf{Theorem} & \textbf{Proof Vector} \\
\hline
a + 0 = a & [1, 0, 0, 0, 1] \\
a + b = b + a & [1, 1, 0, 0, 1] \\
Infinitely Many Primes & [1, 1, 0, 0, 1] \\
\hline
\end{tabular}
\end{center}

\subsection*{ZFC Set Theory (10 dimensions)}
\begin{center}
\begin{tabular}{|l|l|}
\hline
\textbf{Theorem} & \textbf{Proof Vector} \\
\hline
Singleton Set Exists & [1, 1, 0, 0, 0, 0, 0, 0, 0, 0] \\
Union of Two Sets Exists & [1, 0, 1, 1, 0, 0, 0, 0, 0, 0] \\
Power Set Exists & [1, 0, 0, 0, 1, 0, 0, 0, 0, 0] \\
\hline
\end{tabular}
\end{center}

These proof vectors serve as a foundational dataset for building structural maps of mathematical knowledge and enable comparison across domains through a shared logical framework.

\section{Visualization Techniques}

We visualize the logical structure of theorems through heatmaps that display their proof vectors in a matrix form. Each row represents a theorem, and each column corresponds to a specific axiom from a given system. The value in each cell indicates whether the corresponding axiom is used (binary), or to what degree (if using weighted vectors).\\

This visualization allows for immediate comparison of logical profiles across theorems. The use of color gradients in the heatmap highlights similarities and differences in axiom usage, facilitating visual clustering and pattern recognition.\\

Furthermore, similarity between theorems can be quantified using metrics such as cosine similarity, producing a similarity matrix or clustering dendrogram. These methods help identify families of theorems that share common logical foundations, even when they arise from different mathematical domains.

\begin{figure}[h]
\centering
\includegraphics[width=0.9\textwidth]{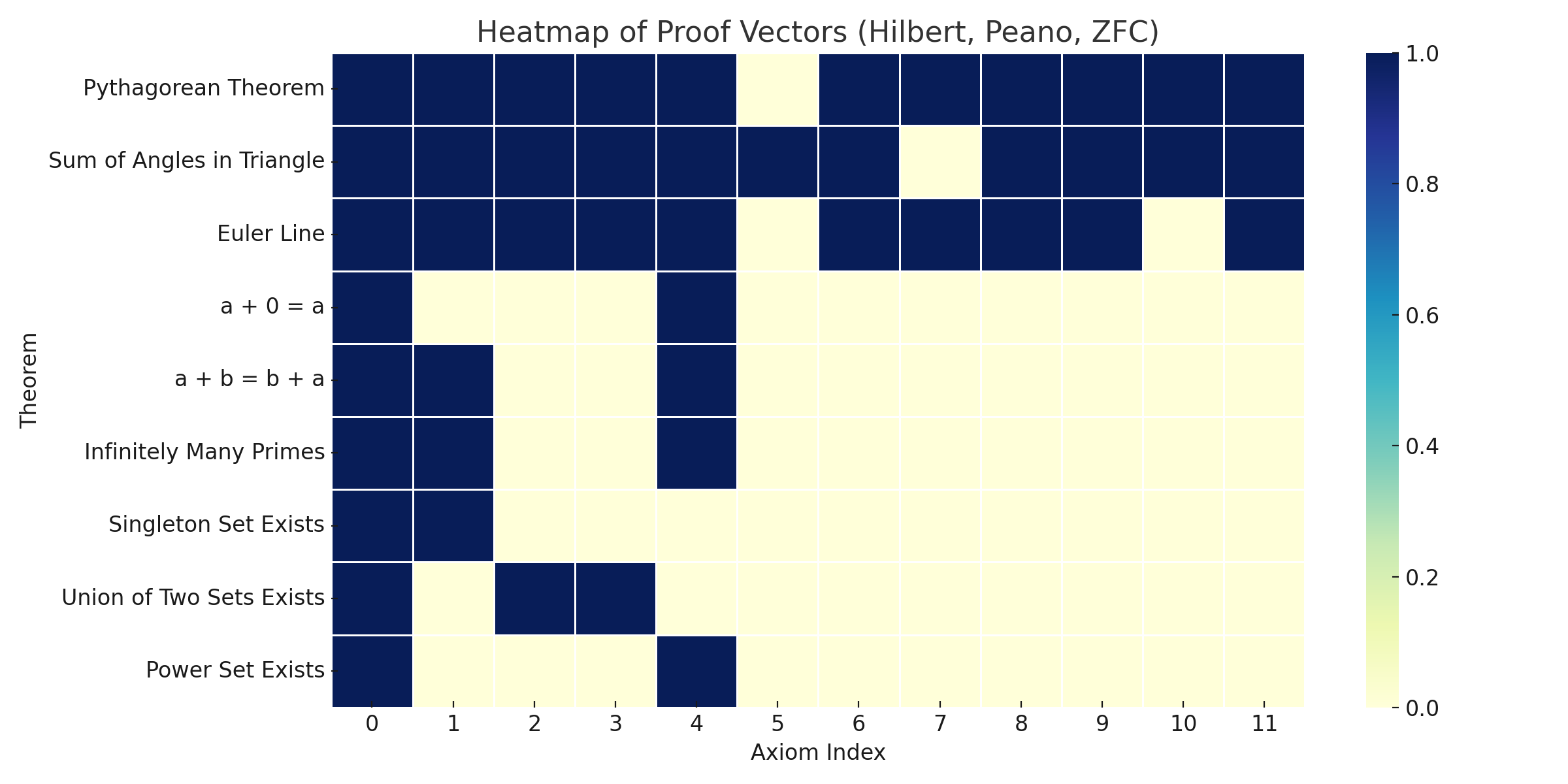}
\caption{Heatmap of proof vectors for selected theorems under Hilbert, Peano, and ZFC axiom systems.}
\label{fig:heatmap}
\end{figure}

\section{Atlas-GPT Assistant: AI Integration}

To facilitate interaction with the Axiom-Based Atlas, we developed a prototype AI assistant, \textbf{Atlas-GPT}, that interprets natural language descriptions of mathematical theorems and suggests corresponding proof vectors. This tool demonstrates how language models can engage with foundational logic in a structured way.

\subsection*{System Architecture}
The assistant consists of the following pipeline.
\begin{enumerate}
  \item \textbf{Input:} A theorem or conjecture described in natural language.
  \item \textbf{Semantic Parsing:} The input is semantically analyzed using a large language model (e.g., GPT-4).
  \item \textbf{Proof Vector Generation:} The system identifies relevant axiom systems and predicts a likely proof vector.
  \item \textbf{Explanation:} A narrative explanation is produced, describing the logical dependencies behind the predicted vector.
  \item \textbf{Output:} The predicted vector, explanation, and related theorems are presented to the user.
\end{enumerate}

\subsection*{Use Case Example}
\textbf{Input:} ``There are infinitely many primes.''\\

\textbf{Output:}
\begin{itemize}
  \item \textbf{Axiom System:} Peano Arithmetic
  \item \textbf{Predicted Proof Vector:} [1, 1, 0, 0, 1]
  \item \textbf{Explanation:} Uses natural number construction and induction.
  \item \textbf{Similar Theorems:} ``There is no largest natural number,'' ``There are infinitely many composites''
\end{itemize}

\noindent
\textbf{Input:} ``The sum of the interior angles of a triangle is 180 degrees.''\\

\textbf{Output:}
\begin{itemize}
	\item \textbf{Axiom System:} Hilbert Geometry Axioms
	\item \textbf{Predicted Proof Vector:} [1, 1, 1, 1, 1, 1, 1, 0, 1, 1, 1, 1]
	\item \textbf{Explanation:} Relies on parallel postulate, congruence and angle addition.
	\item \textbf{Similar Theorems:} ``Pythagorean Theorem'', ``Triangle Similarity Criteria''
\end{itemize}

\subsection*{Future Development}
Atlas-GPT could be extended to provide
\begin{itemize}
  \item Interactive theorem input and editing interface
  \item Integration with formal proof systems (e.g., Lean, Coq)
  \item Training on structured proof corpora
  \item Feedback mechanisms for improving vector prediction accuracy
\end{itemize}

This assistant exemplifies how the Axiom-Based Atlas can bridge natural mathematical language and formal foundational structure, supporting both automated reasoning and educational tools.

\section{Applications and Future Directions}

The Axiom-Based Atlas provides a structural foundation with potential to impact a wide range of areas in mathematics, logic, education, and artificial intelligence. By reframing the way we organize and compare theorems, this framework invites new modes of interaction with mathematical knowledge.

\subsection*{1. Educational Tools and Curriculum Design}
The atlas offers a novel method for students and educators to understand mathematics not only by content but by structure. Instructors can highlight which axioms are at play in a given result and trace dependencies across multiple topics. Proof vectors could also be integrated into textbooks or learning platforms to support concept mapping and personalized learning.

\subsection*{2. Research in Logic and Foundations}
The structural comparison of theorems can offer new insight into the hierarchy and complexity of proofs. Proof vectors allow for formal classification, detection of unexpected commonalities, and exploration of minimal axiom sets needed for certain results. This may inform foundational research in reverse mathematics and proof theory.

\subsection*{3. Mathematical Knowledge Representation}
As mathematics becomes increasingly digitized, the need for structured, searchable, and interoperable representations grows. The atlas serves as a middle ground between fully formal proof assistants and informal natural language databases. It could be used to construct ontologies or semantic networks of mathematics.

\subsection*{4. Integration with Proof Assistants}
The vector-based format can support automated systems by providing training data and logical fingerprints for theorems. Integration with proof assistants such as Coq, Lean, and Isabelle may enable automated generation of minimal proof dependencies, and ultimately support AI-assisted formalization.

\subsection*{5. AI-Augmented Theorem Discovery}
By representing theorems as points in a logical vector space, machine learning models can search for gaps, outliers, or interpolations---possibly suggesting new theorems. Similar to how embeddings drive discovery in NLP and chemistry, proof vector embeddings may unlock new directions in conjecture generation.\\

In future work, we plan to scale the dataset, improve Atlas-GPT’s accuracy, and expand to additional axiom systems and mathematical domains. Ultimately, the goal is to create an interactive, explorable, and extensible atlas of mathematical structure that serves both human understanding and machine reasoning.

\section{Conclusion}

The Axiom-Based Atlas introduces a new paradigm for representing mathematical theorems---not just as standalone results but as structured points within a foundational landscape. By encoding proofs as vectors over formal axiom systems, we enable both qualitative insight and quantitative analysis of mathematical logic. \\

This approach opens the door to new types of visualization, comparison, and discovery, blending human intuition with machine reasoning. The Atlas-GPT prototype further demonstrates the feasibility of integrating natural language, symbolic logic, and AI models into a unified framework.\\

As mathematics continues to expand and digitize, frameworks like the Axiom-Based Atlas offer a scalable method for organizing and interacting with knowledge. Our hope is that this work will lay the groundwork for future research in structural mathematics, AI-assisted reasoning, and formal educational design.

\end{document}